\pdfoutput=1

\documentclass[11pt]{article}

\usepackage[]{acl}

\usepackage{times}
\usepackage{latexsym}

\usepackage{graphicx}
\usepackage{picinpar}
\usepackage{subcaption}
\usepackage{colortbl}
\usepackage{booktabs}
\usepackage{multirow}
\usepackage{soul}
\usepackage{enumerate}
\usepackage{color}
\usepackage{hyperref}
\usepackage{amsmath}
\usepackage{amssymb}
\usepackage{algorithm}
\usepackage{algpseudocode}

\usepackage[T1]{fontenc}

\usepackage[utf8]{inputenc}

\usepackage{microtype}

\title{Domain-specific Continued Pretraining of Language Models \\for Capturing Long Context in Mental Health}

\author{Shaoxiong Ji~\textsuperscript{1} \quad Tianlin Zhang~\textsuperscript{2} \quad Kailai Yang~\textsuperscript{2}\quad Sophia Ananiadou~\textsuperscript{2} \\ \textbf{Erik Cambria~\textsuperscript{3} \quad Jörg Tiedemann~\textsuperscript{1}} \\
	\textsuperscript{1} University of Helsinki \quad
    \textsuperscript{2} The University of Manchester \quad
    \textsuperscript{3} Nanyang Technological University\\
\texttt{~\{shaoxiong.ji;~jorg.tiedemann\}@helsinki.fi;~cambria@ntu.edu.sg} \\
\texttt{\{kailai.yang,tianlin.zhang\}@postgrad.manchester.ac.uk}\\
\texttt{\{sophia.ananiadou\}@manchester.ac.uk}\\
}

\begin{document}
\maketitle
\begin{abstract}
Pretrained language models have been used in various natural language processing applications. 
In the mental health domain, domain-specific language models are pretrained and released, which facilitates the early detection of mental health conditions.
Social posts, e.g., on Reddit, are usually long documents. 
However, there are no domain-specific pretrained models for long-sequence modeling in the mental health domain.
This paper conducts domain-specific continued pretraining to capture the long context for mental health. 
Specifically, we train and release MentalXLNet and MentalLongformer based on XLNet and Longformer.
We evaluate the mental health classification performance and the long-range ability of these two domain-specific pretrained models.
Our models are released in HuggingFace\footnote{\url{https://huggingface.co/AIMH}}.
\end{abstract}

\section{Introduction}
\label{sec:introduction}

Natural Language Processing (NLP) applied to mental healthcare~\citep{le2021machine,zhang2022natural} has received much attention with specific applications to bipolar disorder detection~\citep{harvey2022natural}, depression detection~\citep{ansari-ensemble-hybrid}, and suicidal ideation detection~\citep{ji2021sid}.
Recent work applied pretrained language models and domain-specific continued pretraining in the mental health domain with public models released, such as MentalBERT and MentalRoBERTa~\citep{ji2022mentalbert}.
However, due to the quadratic complexity of self-attention in the transformer network~\citep{vaswani2017attention}, the pretrained Bidirectional Encoder Representations from Transformers (BERT)~\citep{devlin2019bert} and its domain-specific variants have limited ability to capture long-range context, and the pretrianed models can only process sequence within 512 tokens in the downstream applications. 

To address this issue, efficient transformers are proposed, such as Longformer~\citep{beltagy2020longformer} and Transformer-XL~\citep{dai2019transformer} to capture long context. 
\citet{qin2023nlp} conducted a systematic analysis on the long-range ability of efficient transformers. 
In mental healthcare, texts such as self-reported mental conditions are usually long documents. 
For example, in social network analysis on Reddit, users' posts are long, and each user might have multiple posts. 

This paper focuses on mental health analysis with long documents.
We conduct domain-specific continued pretraining with Longformer and XLNet architectures in the mental health domain. 
Our contributions are as follows. We train and release two domain-specific language models, i.e., MentalXLNet and MentalLongformer. 
We evaluate the performance of these two models on various mental healthcare classification datasets.
Finally, we discuss the long-range ability of these two models and summarize how to choose pretrained language representations for specific applications.

\section{Methods and Materials}
\label{sec:methods}

This section presents the methods and materials. 
The self-attention in the standard transformer architecture suffers from quadratic complexity with sequence length. 
As a result, the BERT model pretrained with a masked language modeling (MLM) objective limits the maximum sequence length to 512 tokens.

We introduce two transformer networks for long documents and domain-specific pretraining to continue the pretraining in the mental healthcare domain.
Table~\ref{tab:PTM} summarizes the learning objectives and sequence lengths of existing pretrained models for mental healthcare and models trained in this paper.
For downstream classification tasks, the max sequence length of BERT and RoBERTa is 512. 

\begin{table}[!ht]
\centering
\footnotesize
\begin{tabular}{l | c  c }
\toprule
Model & Objective & Seq. Length \\
\midrule
MentalBERT & MLM & 128 \\
MentalRoBERTa & MLM & 128 \\
MentalXLNet & PLM & 512\\
MentalLongformer & MLM & 4096 \\
\bottomrule
\end{tabular}
\caption{A summary of pretrained models for mental healthcare}
\label{tab:PTM}
\end{table}

\subsection{Transformers for Long Sequence Modeling}
\paragraph{Longformer}
Longformer~\citep{beltagy2020longformer} proposes an efficient attention mechanism with a linear complexity that leverages local windowed attention and task-motivated global attention. 
It is better at autoregressive language modeling on long sequences than prior works. 
When pretrained with MLM objective, Longformer achieves better long sequence modeling capacity on various downstream tasks.

\paragraph{XLNet}
Transformer-XL~\citep{dai2019transformer} solves the context fragmentation issue of fixed-length contexts by devising the recurrent operations for segments in the self-attention network.  
XLNet~\citep{yang2019xlnet} combines the best of autoregressive and autoencoding language modeling and adopts the Permutation Language Modeling (PLM) objective. 
It captures bidirectional context and avoids the discrepancy of masked positions between pretraining and fine-tuning in masked language models.
XLNet has been utilized to train domain-specific models, e.g., Clinical XLNet~\citep{huang2020clinical} in the clinical domain.

\subsection{Domain-specific Continued Pretraining}
We use the same corpus for pretraining as used in MentalBERT~\citep{ji2022mentalbert}.
The pretraining corpus is sourced from Reddit, an online forum of anonymous communities where people with similar interests can engage in discussions. 
For the purpose of our study, we focus on subreddits related to the mental health domain. The selected subreddits for mental health-related content include ``r/depression'', ``r/SuicideWatch'', ``r/Anxiety'', ``r/offmychest'', ``r/bipolar'', ``r/mentalillness'', and ``r/mentalhealth''. 
We obtain the posts of users from these subreddits through web crawling. 
Even though user profiles are publicly available, we do not collect them while gathering the pretraining corpus. 
We conduct continued pretraining on the cluster with Nvidia A100 and V100 GPUs and utilize four GPUs in a single computing node.

\section{Evaluation}
\label{sec:experiments}

Our study aims to investigate the effectiveness of using pretrained language models in detecting binary and multi-class mental disorders such as stress, anxiety, and depression and capturing long context in self-reported mental health texts. 
To achieve this, we fine-tune the language models in downstream tasks.
For the classification model, we adopt the pooled representations of the last hidden layer and the multi-layer perceptron (MLP) with the hyperbolic tangent activation function. 
The learning rate of the transformer text encoder is set to 1e-05, and the learning rate of the classification layers is 3e-05. 
We use the Adam optimizer~\citep{kingma2014adam} for training.

\begin{figure}[!htbp]
\centering
	\includegraphics[width=0.4\textwidth]{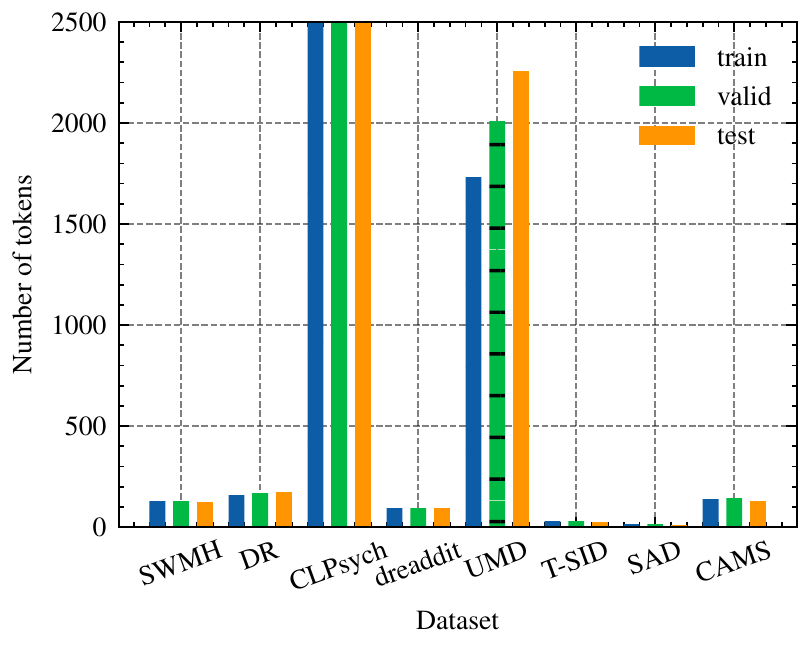}
\caption{Median number of tokens in the datasets}
\label{fig:number_of_tokens}
\end{figure}

\subsection{Datasets}
\label{sec:datasets}

\begin{table*}[!ht]
\footnotesize
\centering
\setlength{\tabcolsep}{6pt}
\begin{tabular}{lllrrr}
\toprule
Category	&	Platform	&	Dataset		&	Train	&	Validation	&	Test	\\
\midrule
Assorted	&	Reddit	&	SWMH	\citep{ji2022suicidal}	&	34,823	&	8,706	&	10,883	\\
Depression	&	Reddit	&	Depression$\_$Reddit	\citep{pirina2018identifying}	&	1,004	&	431	&	406	\\
Depression	&	Reddit	&	CLPsych15	\citep{coppersmith2015clpsych}	&	457	&	197	&	300	\\
Stress	&	Reddit	&	Dreaddit	\citep{turcan2019dreaddit}	&	2,270	&	568	&	715	\\
Suicide	&	Reddit	&	UMD	\citep{shing2018expert}	&	993	&	249	&	490	\\
Suicide	&	Twitter	&	T-SID	\citep{ji2022suicidal}	&	3,072	&	768	&	960	\\
Stress	&	SMS-like	&	SAD	\citep{mauriello2021sad}	&	5,548	&	617	&	685	\\
Assorted & Reddit & CAMS~\citep{garg2022cams} & 2,208 & 321 & 626 \\
\bottomrule
\end{tabular}
\caption{A summary of datasets. Note we hold out a portion of the original training set as the validation set if the original dataset does not contain a validation set.}
\label{tab:data}
\end{table*}%

We consider various mental health classification tasks, i.e., depression, stress, and suicidal ideation, and the cause classification task on stress and mental disorder.
For depression detection, we use two datasets, CLPsych15~\citep{coppersmith2015clpsych} and Depression$\_$Reddit (DR)~\citep{pirina2018identifying}, collected from Reddit. These two datasets contain binary labels. 
For stress, we adopt two datasets, i.e., Dreaddit~\citep{turcan2019dreaddit} and SAD~\citep{mauriello2021sad}, collected from Reddit and short text messages, respectively. 
Dreaddit is a dataset for binary stress classification. The SAD dataset is annotated with stress causes, including school, financial problems, family issues, social relationships, work, health or physical pain, emotional turmoil, everyday decision-making, and other uncategorized causes.
We use the T-SID dataset~\citep{ji2022suicidal} for suicidal ideation detection, which also contains tweets with depression and post-traumatic stress disorders. 

We also use Reddit posts collected by the UMD~\citep{shing2018expert}.
SWMH~\citep{ji2022suicidal} contains posts with several mental health conditions and suicidal ideation. 
The last dataset used in our experiments is CAMS~\citep{garg2022cams} for causal analysis of mental health issues. We utilize the causal categorization, which consists of six categories, i.e., bias or abuse, jobs and career, medication, relationships, alienation, and no reason. 
Datasets are summarized in Table \ref{tab:data} with the median number of tokens of each split shown in Figure \ref{fig:number_of_tokens}. 
The CLPsych and UMD datasets containing multiple posts are extremely long, with over 2,500 and 1,500 tokens in the median, respectively. 
We analyze the long-range ability of MentalXLNet and MentalLongformer on these two datasets.

\subsection{Baselines}
\label{sec:baseline}
We compare our models with three categories of models, i.e., models from the original checkpoint, models with domain-specific continued pretraining, and ChatGPT\footnote{\url{https://openai.com/blog/chatgpt}} in the zero-shot setting. 

\paragraph{Models from the Huggingface Checkpoint}
We consider BERT~\citep{devlin2019bert}, RoBERTa~\citep{liu2019roberta}, XLNet~\citep{yang2019xlnet}, and Longformer~\citep{beltagy2020longformer}, using the model checkpoint released in HuggingFace transformers~\citep{wolf2020transformers}. 
\paragraph{Models with Domain-specific Continued Pretraining}
We then compare our models with two domain-specific models for mental health, i.e., MentalBERT and MentalRoBERTa~\citep{ji2022mentalbert}.

\paragraph{Zero-shot ChatGPT}
We compare our models with zero-shot ChatGPT referring to~\citet{yang-evaluation-chatgpt}.
ChatGPT$_{ZS}$ uses ChatGPT as the inference engine with simple manual prompts,
ChatGPT$_{V}$, ChatGPT$_{N\_sen}$, and ChatGPT$_{N\_emo}$ inject distant supervision signals from lexicons, i.e., NRC EmoLex sentiments~\citep{Mohammad13emolex}, VADER sentiments~\citep{hutto2014vader}, and SenticNet~\citep{camnt7}, to the prompts. 
ChatGPT$_{CoT}$ and ChatGPT$_{CoT\_emo}$ utilize the Chain-of-Thought method~\citep{wei2022chain} and its enhancement with emotion~\citep{amiwil}.

\subsection{Main Results}

Table~\ref{tab:results} presents the results on eight test sets, comparing the models trained in this paper with baselines in a supervised setting and with ChatGPT in the zero-shot setting. 
Table~\ref{tab:results_witout_chatgpt} only compares the performance with supervised baselines because ChatGPT does not achieve strong performance compared to strong supervised methods, and we have no budget to run more experiments with ChatGPT. 
MentalXLNet and MentalLongformer outperform other baselines in most cases, especially on datasets with longer sequences, such as CLPsych15. 
MentalRoBERTa performs best on Dreaddit, showing its robust performance on short sequences, while MentalLongformer also gets a comparable performance. 
On the SMS-like SAD dataset, Longformer achieves the best performance, while the MentalLongformer trained on Reddit gets a slightly worse performance. 

\begin{table*}[!hbt]
\footnotesize
\centering
\begin{tabular}{l|cc|cc|cc|cc|cc|cc}
\toprule
\multicolumn{1}{c}{\multirow{2}{*}{\textbf{Model}}} & \multicolumn{2}{c}{\textbf{DR}} & \multicolumn{2}{c}{\textbf{CLPsych15}} & \multicolumn{2}{c}{\textbf{Dreaddit}} & \multicolumn{2}{c}{\textbf{T-SID}} & \multicolumn{2}{c}{\textbf{SAD}} & \multicolumn{2}{c}{\textbf{CAMS}} \\
\multicolumn{1}{c|}{} & Rec. & F1 & Rec. & F1 & Rec. & F1 & Rec. & F1 & Rec. & F1 & Rec. & F1 \\ \midrule
BERT & 91.13 & 90.90 & 64.67 & 62.75 & 78.46 & 78.26 & 88.44 & 88.51 & 62.77 & 62.72 & 40.26 & 34.92 \\
RoBERTa & {95.07} & {95.11} & 67.67 & 66.07 & 80.56 & 80.56 & 88.75 & 88.76 & 66.86 & 67.53 & 41.18 & 36.54 \\
XLNet	&	90.89	&	90.44	&	69.83	&	69.12	&	78.88	&	78.84	&	86.04	&	86.18	&	67.30	&	67.30	&	50.64	&	49.16	\\
Longformer	&	95.81	&	95.74	&	75.67	&	75.47	&	81.54	&	81.45	&	89.58	&	89.63	&	\textbf{69.20}	&	\textbf{69.01}	&	49.52	&	49.42	\\
\midrule
MentalBERT & 94.58 & 94.62 & 64.67 & 62.63 & 80.28 & 80.04 & 88.65 & 88.61 & 67.45 & 67.34 & 45.69 & 39.73 \\
MentalRoBERTa & 94.33 & 94.23 & {70.33} & {69.71} & \textbf{81.82} & \textbf{81.76} & {88.96} & {89.01} & {68.61} & {68.44} & {50.48} & {47.62} \\ 
\midrule
ChatGPT$_{ZS}$ & 82.76 & 82.41 & 60.33 & 56.31 & 72.72 & 71.79 & 39.79 & 33.30 & 55.91 & 54.05 & 32.43 & 33.85 \\
ChatGPT$_{V}$ & 79.51 & 78.01 & 59.20 & 56.34 & 74.23 & 73.99 & {40.04} & {33.38} & 52.49 & 50.29 & 28.48 & 29.00 \\
ChatGPT$_{N\_sen}$ & 80.00 & 78.86 & 58.19 & 55.50 & 70.87 & 70.21 & 39.00 & 32.02 & 52.92 & 51.38 & 26.88 & 27.22 \\
ChatGPT$_{N\_emo}$ & 79.51 & 78.41 & 58.19 & 53.87 & 73.25 & 73.08 & 39.00 & 32.25 & 54.82 & 52.57 & 35.20 & 35.11 \\
ChatGPT$_{CoT}$ & 82.72 & 82.90 & 56.19 & 50.47 & 70.97 & 70.87 & 37.66 & 32.89 & 55.18 & 52.92 & 39.19 & 38.76 \\
ChatGPT$_{CoT\_emo}$ &{ 83.17} & {83.10} & {61.41} & {58.24} & {75.07} & {74.83} & 34.76 & 27.71  & {58.31} & {56.68} & {43.11} & {42.29} \\
\midrule
MentalXLNet	&	95.32	&	95.24	&	{71.67}	&	{71.49}	&	80.42	&	80.41	&	89.17	&	89.12	&	\textbf{69.20}	&	68.76	&	\textbf{50.80}	&	\textbf{50.08}	\\
MentalLongformer	&	\textbf{96.55}	&	\textbf{96.53}	&	\textbf{77.00}	&	\textbf{76.32}	&	81.12	&	81.05	&	\textbf{89.90}	&	\textbf{89.89}	&	68.76	&	68.44	&	49.20	&	48.74	\\
\bottomrule
\end{tabular}
\caption{Results of mental health classification. The bold text represents the best performance. Note that: for Longformer and MentalLongformer, the best results are reported with longer texts as inputs.}
\label{tab:results}
\end{table*}

\begin{table}[!ht]
\footnotesize
\centering
\begin{tabular}{l|cc|cc}
\toprule
\multicolumn{1}{c}{\multirow{2}{*}{\textbf{Model}}} & \multicolumn{2}{c}{\textbf{UMD}} & \multicolumn{2}{c}{\textbf{SWMH}} \\
\multicolumn{1}{c|}{} & Rec. & F1 & Rec. & F1 \\
\midrule
BERT	&	61.63	&	58.01	&	69.78	&	70.46	\\
RoBERTa	&	59.39	&	60.26	&	70.89	&	72.03	\\
XLNet	&	63.06	&	60.09	&	70.60	&	70.57	\\
Longformer	&	\textbf{74.29}	& \textbf{72.85}	&	71.86	&	71.79	\\
\midrule
MentalBERT	&	64.08	&	58.26	&	69.87	&	71.11	\\
MentalRoBERTa	&	57.96	&	58.58	&	70.65	&	\textbf{72.16}	\\
\midrule
MentalXLNet	&	63.06	&	{60.29}	&	71.07	&	71.18	\\
MentalLongformer	&	73.06	&	72.47	&	\textbf{72.05}	&	72.08	\\
\bottomrule
\end{tabular}
\caption{Results on mental health classification on UMD and SWMH}
\label{tab:results_witout_chatgpt}
\end{table}

\subsection{Results of Long-range Ability }

We analyze the long-range ability by inputting various lengths of text into the model. 
Table~\ref{tab:long-range} shows the results on UMD and CLPsych15 datasets with longer documents than others.
The results indicate an increasing trend in recall and F1 scores with a certain degree of fluctuation when the sequence length increases. 
Domain-specific continued pretraining continues to improve performance in most cases. 

These results show the long-range ability of Longformer and XLNet and their domain-specific variants and also verify the effectiveness of domain-specific continued pretraining.
However, there is no clear explanation for the fluctuation due to the black-box nature of transformers and the lack of human-grounded evaluation. 
One possible guess is that longer texts provide more information but can also introduce redundancy that impairs the model performance. 

\begin{table}[!ht]
\footnotesize
\centering
\begin{tabular}{l| l|cc|cc}
\toprule
\multirow{2}{*}{\textbf{Dataset}} & \textbf{Seq.} & \multicolumn{2}{c}{\textbf{Longformer}} & \multicolumn{2}{c}{\textbf{MentalLongformer}} \\
& \textbf{Len.} & Rec. & F1 & Rec. & F1 \\
\midrule
\multirow{8}{*}{UMD} & 512	&	65.10	&	58.36	&	62.24	&	\textbf{59.74}	\\
& 1024	&	63.27	&	62.34	&	\textbf{64.29}	&	\textbf{66.22}	\\
& 1536	&	67.55	&	66.90	&	67.55	&	66.90	\\
& 2048	&	65.92	&	67.90	&	\textbf{70.82}	&	\textbf{69.19}	\\
& 2560	&	68.98	&	68.10	&	\textbf{72.04}	&	\textbf{72.53}	\\
& 3072	&	71.43	&	72.15	&	\textbf{72.65}	&	69.76	\\
& 3584	&	62.65	&	66.08	&	\textbf{72.45}	&	\textbf{72.13}	\\
& 4096	&	74.29	&	72.85	&	73.06	&	72.47	\\
\midrule
\multirow{8}{*}{CLP} & 512	&	64.33	&	63.44	&	59.00	&	54.85	\\
& 1024	&	70.67	&	69.68	&	\textbf{71.33}	&	\textbf{70.76}	\\
& 1536	&	69.00	&	67.27	&	\textbf{71.00}	&	\textbf{69.57}	\\
& 2048	&	75.33	&	75.26	&	\textbf{72.32}	&	\textbf{72.00}	\\
& 2560	&	75.00	&	74.57	&	\textbf{76.00}	&	\textbf{75.69}	\\
& 3072	&	65.33	&	62.53	&	\textbf{72.33}	&	\textbf{70.97}	\\
& 3584	&	72.00	&	70.91	&	\textbf{75.00}	&	\textbf{74.31}	\\
& 4096	&	75.67	&	75.47	&	\textbf{77.00}	&	\textbf{76.32}	\\
\bottomrule
\end{tabular}
\caption{Long-range ability analysis on UMD and CLPsych15. The bold text indicates that MentalLongformer achieves better scores.}
\label{tab:long-range}
\end{table}

\section{Related Work}
\label{sec:related}

Mental health surveillance in social media has gained increasing research attention from the NLP community.
\citet{le2021machine} conducted a systematic review of machine learning and NLP in mental health, summarized the state of the art in this field, discussed the challenges and opportunities, and provided recommendations for future research.

Emotion information in social posts is an important cue for mental health detection. 
\citet{zhang-emotion-fusion-survey} surveyed emotion fusion methods for mental illness detection from social media. 
Various approaches used NLP and machine learning for mental health classification. 
\citet{krishnamurthy2016hybrid} presented a hybrid statistical and semantic model for identifying mental health and behavioral disorders using social network analysis. 
\citet{ji2018supervised} evaluated supervised learning methods for detecting suicidal ideation in online user content. 
\citet{nijhawan2022stress} studied stress detection.
\citet{spruit2022exploring} explored language markers of mental health in psychiatric stories. 

These papers highlight the potential of natural language processing and machine learning for improving the early detection of mental health conditions. 
\citet{ive2020generation} focused on the generation and evaluation of artificial mental health records for natural language processing.
\citet{ji-towards-intention-understanding} emphasized the importance of intention understanding in suicide risk assessment with pretrained language models.

\section{Conclusion}
\label{sec:conclusion}
We train and release two domain-specific language models in mental health, i.e., MentalXLNet and MentalLongformer.
We empirically analyze the performance of these two models on various mental health classification datasets. 
We validate that the domain-specificity of pretrained language models can improve the performance of downstream tasks.
For short texts within 512 tokens, we recommend MentalRoBERTa and MentalXLNet. 
For longer texts, MentalLongformer is a better choice. 

\section*{Ethical Statement}
Privacy is important in the mental health domain. 
We use social media posts that are manifestly public and do not collect user profiles when pretraining language models and fine-tuning classification models. 
We use the corpus collected from Reddit and do not interact with users who post on Reddit. 
Language models can be biased if the collected social posts contain inherent biases.
Models pretrained and fine-tuned in this paper can not replace psychiatric diagnoses. 
We recommend individuals experiencing a mental health condition seek help from mental health professionals.

\section*{Acknowledgments}
The authors wish to acknowledge CSC – IT Center for Science, Finland, for computational resources.

\section*{Limitations}
We conduct domain-specific continued pretraining on a Reddit corpus, which means the domain-specificity primarily applies to Reddit social media. 
Social media data can be biased. 
However, we did not investigate this issue due to the lack of resources. 
This could be a future research task.

\bibliography{references}
\bibliographystyle{acl_natbib}

\end{document}